\documentclass[letterpaper, 10 pt, conference]{ieeeconf}  

\IEEEoverridecommandlockouts                              

\usepackage{amsmath}
\usepackage{amssymb}
\usepackage{graphicx}
\usepackage{caption}
\usepackage{subcaption}
\usepackage{tabularx,booktabs}
\usepackage{bm}
\usepackage{multirow}
\usepackage[table,xcdraw]{xcolor}

\title{\LARGE \bf
Tactile Functasets: Neural Implicit Representations of Tactile Datasets
}

\author{Sikai Li, Samanta Rodriguez, Yiming Dou, Andrew Owens, Nima Fazeli
\thanks{*Supported by NSF GRFP \#2241144, NSF CAREER Awards \#2339071 and \#2337870, and NSF NRI \#2220876.}
\thanks{University of Michigan, Ann Arbor, MI, USA
        {\tt\small <skevinci,samanrod,ymdou,ahowens,nfz>@umich.edu}}%
}

\renewcommand{\vec}{\bm}
\newcommand{\mat}[1]{\bm{#1}}

\begin{document}

\maketitle
\thispagestyle{empty}
\pagestyle{empty}

\begin{abstract}
    Modern incarnations of tactile sensors produce high-dimensional raw sensory feedback such as images, making it challenging to efficiently store, process, and generalize across sensors. To address these concerns, we introduce a novel implicit function representation for tactile sensor feedback. Rather than directly using raw tactile images, we propose neural implicit functions trained to reconstruct the tactile dataset, producing compact representations that capture the underlying structure of the sensory inputs. These representations offer several advantages over their raw counterparts: they are compact, enable probabilistically interpretable inference, and facilitate generalization across different sensors. We demonstrate the efficacy of this representation on the downstream task of in-hand object pose estimation, achieving improved performance over image-based methods while simplifying downstream models. We release code, demos and datasets at https://www.mmintlab.com/tactile-functasets.
\end{abstract}

\captionsetup{font=small}
\section{INTRODUCTION}

Tactile sensing plays a crucial role in enabling dexterous manipulation, providing rich feedback from the physical interaction between the robot and its environment. However, unlike more standardized sensing modalities like vision, tactile sensing has not yet reached the same level of widespread standardization, resulting in a diverse range of sensor designs and data formats. Many modern tactile sensors, such as GelSight~\cite{yuan2017gelsight,johnson2009retrographic}, Soft Bubble~\cite{softbub_tedrake}, GelSlim~\cite{gelsim_donlon}, Finger Vision~\cite{fingervision}, DIGIT~\cite{lambeta2020digit}, and  DenseTact~\cite{Do2022DenseTactOT} output raw, high-dimensional data in the form of images or pixel grids. This raw data format poses several challenges for downstream tasks, as it is computationally expensive to process and store, and difficult to generalize across different sensor designs. 

Current approaches to using tactile sensing for downstream applications largely rely on either direct usage of raw data or hand-engineered features extracted from tactile images. While these methods have shown success in specific applications, they come with several limitations. First, raw image data from tactile sensors can be high-dimensional, making real-time inference slow and resource-intensive. Second, models trained on data from one sensor often do not generalize well to others, requiring new models and algorithms to be developed for each sensor type. Finally, image-based representations can be cumbersome to incorporate into probabilistic frameworks for tasks that benefit from uncertainty estimation, such as object pose estimation or contact state classification. These limitations hinder the wider adoption of tactile sensing in robotics, especially in scenarios that demand real-time performance and sensor flexibility.

\begin{figure}[t]
    \centering
    \includegraphics[width=0.85\columnwidth]{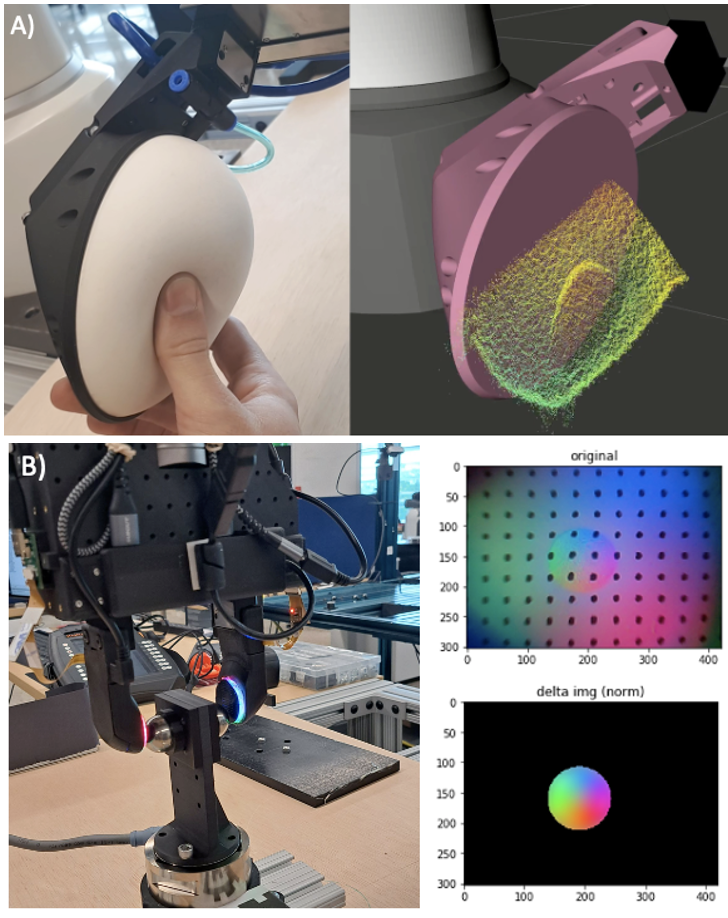}
    \caption{Modern tactile sensors are capable of streaming high-resolution feedback from the contact between the robot and objects. These high-resolution data streams pose significant challenges for processing and storage. We ask whether there is a more compact and efficient method to store data and perform inference for tactile data? This question is inspired by the observation that tactile signals are lower in information density when compared to natural image.}
    \label{fig:teaser}
\end{figure}

In this paper, we propose a novel approach to tactile sensor feedback representation using neural implicit functions. Instead of using raw tactile images, we train neural networks to reconstruct the tactile dataset, resulting in compact and expressive representations. These implicit functions capture the underlying structure of the tactile data and offer several advantages over traditional representations. First, the compactness of the neural representations reduces computational burden, making them more suitable for real-time inference. Second, the flexibility of these neural networks allows for probabilistically interpretable inference strategies, which are essential for tasks that require uncertainty handling. Finally, this representation lend themselves favorably to generalization across different sensor types, enabling seamless transfer of models between sensors without extensive retraining.

We demonstrate the effectiveness of our method by applying it to the downstream task of in-hand object pose estimation, a fundamental challenge in robotic manipulation. Our experiments show that the neural representations not only achieve improved performance over traditional image-based methods but also simplify downstream models used for inference. By leveraging neural implicit functions, we pave the way for more efficient and scalable tactile sensing strategies that can be readily adopted across a variety of robotic platforms and sensor configurations. This work sets the stage for future research in generalizable tactile perception and opens up new possibilities for tactile-driven manipulation in complex environments.
\section{RELATED WORK}

\noindent\textbf{Visuo-tactile Sensors:} Over the past decade, the robotics community has embraced a range of vision-based tactile sensors, including GelSight~\cite{yuan2017gelsight,johnson2009retrographic}, Soft Bubble~\cite{softbub_tedrake}, GelSlim~\cite{gelsim_donlon}, Finger Vision~\cite{fingervision}, DIGIT~\cite{lambeta2020digit}, and DenseTact~\cite{Do2022DenseTactOT}. These sensors convert tactile inputs into high-dimensional visual data, typically represented as 2D images or 3D point clouds. While these rich representations have significantly enhanced performance in robotic manipulation tasks, they often result in large, high-dimensional datasets that are costly to store and process. In certain tasks, visuo-tactile sensors have proven indispensable~\cite{calandra2018more,digit_tactile_sensor,oller2023manipulation,li2014localization,kim2022active}. For our experiments, we employ the Soft Bubble and GelSlim 3.0 sensors. The Soft Bubble sensor~\cite{softbub_tedrake} is made of a compliant, air-filled membrane paired with a depth camera. It perceives tactile data through the membrane's deformations when contact occurs, captured using a time-of-flight sensor. GelSlim~\cite{taylor2022gelslim}, on the other hand, measures tactile signals by capturing deformation in an elastomer with opaque skin, illuminated by LEDs. These sensors differ significantly in contact area, compliance, and output data, ranging from 2D to 3D representations. While these rich signals offer detailed information about the contact surface, much of this data is redundant, leading to unnecessary storage overhead.

Our approach addresses these challenges by introducing a method that not only reduces the storage burden but also enables probabilistic inference, allowing us to infer the tactile signals' underlying structure without relying on the full high-dimensional data.

\noindent\textbf{Tasks and Algorithms for Tactile Sensors:}
Most current tactile sensing models are tailored to specific sensors, leveraging their unique ability to capture local geometry, force, or texture~\cite{yang2023generating,li2014localization}. These sensor-specific representations, while effective, limit generalization across different platforms and are often tied to their respective data formats. In robotics, sensor-specific methods have been developed for tasks like in-hand object pose estimation (e.g., Soft Bubble~\cite{kuppuswamy2019fast}, GelSlim~\cite{kim2022active}, DIGIT~\cite{suresh2023midastouch}), local geometry estimation~\cite{taylor2022gelslim}, and force field estimation across contact surfaces (e.g., Finger Vision~\cite{yamaguchi2016combining}). These techniques have enabled successful performance on various tasks such as peg-in-hole insertion~\cite{kim2022active}, in-hand manipulation~\cite{oller2023manipulation}, and dense packing~\cite{ai2024robopack}. However, their reliance on high-dimensional, sensor-specific data makes them inefficient in terms of storage and generalization.

Our method mitigates these inefficiencies by proposing a transformation framework that reduces the need to store entire high-dimensional tactile images. Instead, we learn compact neural representations of the tactile data, which not only reduce storage costs but also facilitate cross-modal transfer and probabilistic inference, paving the way for sensor-agnostic manipulation strategies.

\section{METHOD}

This section describes our approach to (i) representing tactile sensor data using neural implicit functions, (ii) constructing a tactile functaset, (iii) performing inference over this functaset, and (iv) training downstream models for object pose estimation. 

\subsection{Implicit Function Representation Learning}

We formulate the problem of learning tactile representations as training a neural implicit function to reconstruct tactile sensor data. A neural implicit function in this context is a continuous learnable function $f_\theta : \mathcal{X} \rightarrow \mathcal{F}$ that maps spatial coordinates $x \in \mathcal{X}$ to pixel values $y \in \mathcal{Y}$ of raw tactile images. This function is parameterized by a feedforward neural network with weights $\theta$ and uses sinusoidal activation functions~\cite{sitzmann2020implicitneuralrepresentationsperiodic}, which allow the model to capture high-frequency details.

Given a tactile image's pixels $\mathcal{T} = {(\vec{x}_i, \vec{y}_i)}_{i=1}^N$, where $\vec{x}_i \in \mathbb{R}^2$ represents the pixel location and $\vec{y}_i$ is the corresponding pixel value, we aim to minimize the reconstruction error between the predicted and true pixel values across the image. The optimization objective is:
\begin{equation} 
    \vec{\theta}^* = \operatorname{argmin}_{\vec{\theta}} \sum_{i=1}^N \mathcal{L}(f_{\vec{\theta}} (\vec{x}_i), \vec{y}_i) 
\end{equation}
where $\mathcal{L}$ is the loss function, typically the mean squared error (MSE), which quantifies the difference between the predicted pixel values and the ground truth. This general formulation allows each neural network to encode a single tactile image. However, in practical scenarios, training one network per image can be computationally expensive. In the following subsection, we discuss how we modify this approach to enable a single trunk network with modulations, reducing the need to learn a separate model for each image.

\subsection{Tactile Functaset}

To address the inefficiency of training separate models for each data point, we introduce the concept of a tactile functaset, which captures the variability across multiple tactile inputs using a shared base network. Inspired by \cite{dupont2022datafunctadatapoint}, we leverage a shared trunk network architecture, specifically using SIREN~\cite{sitzmann2020implicitneuralrepresentationsperiodic}, to learn common features across all tactile data. This trunk network encodes the global structure of the tactile signals while allowing specific data points to be represented through modulations. The latent modulation vectors are compact vectors that modulate the trunk network by modifying the network weights in order to capture data-specific variations~\cite{chan2021piganperiodicimplicitgenerative}. By using this approach, the functaset stores only the modulation vectors for each tactile sample, significantly reducing the memory and computational costs. 

In detail, let $\mathcal{D} = {(\mathcal{T}_1, \mathcal{T}_2, \dots, \mathcal{T}_M)}$ denote a tactile dataset consisting of $M$ tactile images, where each $\mathcal{T}_i = {(\vec{x}_{i,j}, \vec{y}_{i,j})}_{j=1}^{N_i}$ represents the $N_i$ pixel-value pairs $(\vec{x}_{i,j}, \vec{y}_{i,j})$ for the $i$-th tactile image. Our goal is to generate a compact functaset that encodes all the tactile signals in $\mathcal{D}$ by parameterizing them as neural functions with shared and data-specific components.

We define a trunk network $f_{\vec{\theta}} : \mathcal{X} \to \mathcal{F}$ with parameters $\theta$, shared across the entire dataset, that captures the common structure present in the tactile signals. The trunk network learns a base mapping from the spatial coordinates $\vec{x}_{i,j} \in \mathbb{R}^2$ to an intermediate feature space $\mathcal{F}$. However, to account for variations between individual tactile signals, we introduce a latent modulation vector $\vec{z}_i \in \mathbb{R}^d$ for each tactile image $\mathcal{T}_i$. The modulation vector $\vec{z}_i$ adjusts the trunk network’s output to represent the specific tactile image.

\begin{figure}[t]
    \centering
    \includegraphics[width=\columnwidth]{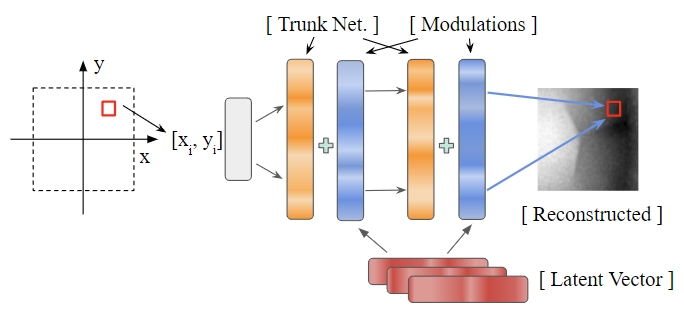}
    \caption{The architecture of our tactile functaset framework. Orange layers represent the trunk model, with blue layers being the modulations mapped from red latent vectors. Each red latent vector is stored as one tactile functa point for a specific data point. The entire model takes pixel locations of tactile images as input and output corresponding pixel values.}
    \label{fig:tactfunc_arch}
\end{figure}

The modulated neural function is thus formulated as $ f_{\vec{\theta}}(\vec{x}; \; \vec{z}_i)$ where the modulation $\vec{z}_i$ is used to condition the trunk network $f_{\theta}$, effectively customizing the network for each tactile image $\mathcal{T}_i$. This conditioning is achieved by applying a linear transformation to the latent vector: 
\begin{align}
    \vec{z}_i \; \rightarrow \; \gamma_i = \mat{W} \vec{z}_i
\end{align}
where $\mat{W} \in \mathbb{R}^{d \times d}$ is a learned transformation matrix that outputs the modulation parameters $\gamma_i$ used to adjust the trunk network at various layers. The latent vector $\vec{z}_i$ provides a compact representation for each image, which is much smaller in size than the full parameterization of the network, making it efficient for storage and retrieval.

The trunk network $f_{\theta}$ and the modulation vectors $\vec{z}_i$ are learned jointly in a meta-learning framework. Specifically, we adopt a bi-level optimization approach, where the trunk network is updated over the entire dataset, and the modulation vectors are learned for each individual sample. The overall loss function for optimizing the parameters of the trunk network and modulations is:
\begin{align}
    \mathcal{L}(\theta, \{\vec{z}_i\}) = \frac{1}{M} \sum_{i=1}^{M} \sum_{j=1}^{N_i} \mathcal{L}_{\text{rec}}(f_{\theta, \vec{z}_i}(\vec{x}_{i,j}), \vec{y}_{i,j})
\end{align}
where $\mathcal{L}_\text{rec}$ denotes the reconstruction loss (e.g., mean squared error) between the predicted and actual pixel values. The reconstruction loss encourages the model to accurately predict the tactile signals for each image in the dataset while maintaining a compact and shared representation. This approach is similar to 
the method described in~\cite{dupont2022datafunctadatapoint}. 

\subsection{Inference over Tactile Functasets}

The inference process involves retrieving the neural implicit representation for a tactile signal from the functaset. Given a new tactile input, we aim to optimize its latent modulation vector while keeping the trunk network fixed. The spatial coordinates of the new tactile signal are fed into the trunk network, which is modulated by the corresponding latent vector to output the reconstructed pixel values.

Let $\vec{x}_{\text{new}}$ be the spatial coordinates of the new tactile signal and $\vec{y}_{\text{new}}$ its corresponding pixel values. The objective is to find the optimal modulation vector $\vec{z}_{\text{new}}$ that minimizes the reconstruction loss, while the trunk network $f_\theta$ remains unchanged. Formally, the optimization problem can be expressed as:
\begin{align}
\vec{z}_{\text{new}}^* = \operatorname{argmin}_{\vec{z}_{\text{new}}} \sum_{j=1}^{N_{\text{new}}} \mathcal{L}_{\text{rec}}(f_{\theta, \vec{z}_{\text{new}}}(\vec{x}_{\text{new}, j}), \vec{y}_{\text{new}, j}),
\end{align}
where $N_{\text{new}}$ is the number of pixel locations in the new tactile signal, and $\mathcal{L}_{\text{rec}}$ is the reconstruction loss (e.g., MSE).

During inference, $\vec{z}_{\text{new}}$ is initialized randomly, and gradient descent is used to update $\vec{z}_{\text{new}}$ iteratively. The update rule for the latent modulation vector at each step $t$ is given by:
\begin{align}
    \vec{z}_{\text{new}}^{t+1} = \vec{z}_{\text{new}}^t - \eta \nabla_{\vec{z}_{\text{new}}} \sum_{j=1}^{N_{\text{new}}} \mathcal{L}_{\text{rec}}(f_{\theta, \vec{z}_{\text{new}}^t}(\vec{x}_{\text{new}, j}), \vec{y}_{\text{new}, j}),
\end{align}
where $\eta$ is the learning rate, and $\nabla_{\vec{z}_{\text{new}}}$ represents the gradient with respect to the latent modulation vector. The optimization continues until convergence, i.e., when the reconstruction error stabilizes or the maximum number of iterations is reached. The final modulation vector $\vec{z}_{\text{new}}^*$ captures the essential features of the new tactile input in the functaset, allowing the trunk network to generate the corresponding tactile image with minimal error. To improve the robustness of inference, we augment our proposed scheme with two techniques: Stochastic Gradient Langevin Dynamics (SGLD) and $k$-Nearest Neighbors ($k$-NN).

\vspace{3pt}
\noindent\textbf{Stochastic Gradient Langevin Dynamics (SGLD):} SGLD provides a way to sample from the posterior distribution over the latent modulation vector $\vec{z}_{\text{new}}$, rather than only optimizing for a point estimate. By introducing controlled noise into the gradient updates, SGLD can produce multiple samples from the posterior, allowing for more uncertainty-aware predictions. The update rule for $\vec{z}_{\text{new}}$ under SGLD is given by:
\begin{align*}
\vec{z}_{\text{new}}^{t+1} &= \vec{z}_{\text{new}}^t - \eta \nabla_{\vec{z}_{\text{new}}} \sum_{j=1}^{N_{\text{new}}} \mathcal{L}_{\text{rec}}(f_{\theta, \vec{z}_{\text{new}}^t}(\vec{x}_{\text{new}, j}), \vec{y}_{\text{new}, j}) \nonumber \\
&\quad + \sqrt{2 \eta} \, \xi^t,
\end{align*}
where $\xi^t \sim \mathcal{N}(0, I)$ is Gaussian noise and $\eta$ is the learning rate. The added noise term $\sqrt{2 \eta} \, \xi^t$ ensures that the modulation vector is updated in a way that reflects the underlying posterior distribution, with the noise magnitude scaled by the learning rate $\eta$. 
After running multiple iterations of SGLD, we obtain a set of samples $\{ \vec{z}_{\text{new}}^{(s)} \}_{s=1}^S$, where $S$ denotes the total number of samples. These samples can then be used for downstream tasks, for instance, by taking an ensemble or by estimating the uncertainty in the latent space.

\vspace{3pt}
\noindent\textbf{$k$-Nearest Neighbors ($k$-NN).} 
As an alternative to SGLD, we propose a simpler inference mechanism based on $k$-NN. Since the trunk network encodes shared features across the entire dataset, similar tactile signals will have similar latent modulation vectors. We can exploit this property by applying $k$-Nearest Neighbors ($k$-NN) to find the closest latent vectors in the functaset to the new input. 

Let $\mathcal{Z} = \{ \vec{z}_i \}_{i=1}^M$ represent the set of latent modulation vectors in the functaset, where $M$ is the total number of samples in the dataset. For a new tactile input, we find the $k$ nearest latent vectors from $\mathcal{Z}$ based on a distance metric $d$, typically Euclidean distance in the latent space:
\begin{align*}
\mathcal{N}(\vec{z}_{\text{new}}) &= \operatorname{argmin}_{\vec{z}_i \in \mathcal{Z}} d(\vec{z}_{\text{new}}, \vec{z}_i), \quad \text{for } i = 1, \dots, k.
\end{align*}
These $k$ nearest neighbors can then be used to approximate the posterior over $\vec{z}_{\text{new}}$ by interpolating between them. For example, the latent vector can be expressed as a weighted combination of the $k$ nearest neighbors:
\begin{align}
\vec{z}_{\text{new}} &\approx \sum_{i=1}^k w_i \vec{z}_i,
\end{align}
where the weights $w_i$ are computed based on the inverse distance. This $k$-NN-based augmentation allows us to make fast inferences by leveraging previously computed modulations, reducing the reliance on gradient-based optimization, and improving the overall inference speed while maintaining accuracy. Additionally, combining $k$-NN with SGLD sampling provides a mechanism to initialize the latent modulation vector $\vec{z}_{\text{new}}$ close to known solutions, which can lead to faster convergence in the optimization process.

\subsection{Downstream Model Learning}

Once the tactile data is represented as a set of neural implicit functions, we train downstream models to predict task-specific outputs, such as in-hand object pose estimation. Let $\vec{z}_i$ represent the latent modulation vector for the $i$-th tactile sample in the functaset, with the corresponding neural implicit representation $f_{\theta, \vec{z}_i}$. The input to the downstream model is the set of neural representations generated by the trunk and modulations, rather than the raw tactile signals.

For in-hand object pose estimation, we aim to predict the object's pose, denoted by $\vec{p}_i$, given the neural representation of the tactile data. The problem can be formulated as learning a function $g_\phi : \mathcal{Z} \rightarrow \mathcal{P}$, parameterized by a downstream model with weights $\phi$, where $\mathcal{Z}$ is the space of latent modulation vectors and $\mathcal{P}$ is the space of object poses:
\begin{align}
\hat{\vec{p}}_i &= g_\phi(\vec{z}_i),
\end{align}
where $\hat{\vec{p}}_i$ is the predicted pose of the object, and $\vec{z}_i$ is the latent modulation vector associated with the $i$-th tactile signal. The model is trained using the loss function given by the discrepancy between the predicted pose $\hat{\vec{p}}_i$ and the ground truth pose $\vec{p}_i$ for each training sample:
\begin{align}
\mathcal{L}_{\text{pose}}(\phi) &= \frac{1}{M} \sum_{i=1}^M \mathcal{L}_{\text{reg}}(g_\phi(\vec{z}_i), \vec{p}_i),
\end{align}
where $M$ is the total number of training samples and $\mathcal{L}_{\text{reg}}$ is a regression loss, such as the mean squared error (MSE):
\begin{align}
\mathcal{L}_{\text{reg}}(\hat{\vec{p}}_i, \vec{p}_i) &= \|\hat{\vec{p}}_i - \vec{p}_i\|^2.
\end{align}
By training the downstream model $g_\phi$ on the functaset-based representations, we exploit the compact and structured information encoded in the neural implicit functions. This significantly reduces the dimensionality of the input space and simplifies the complexity of the model architecture compared to approaches that rely on raw tactile image inputs.

\section{EXPERIMENTS AND RESULTS}
We evaluate our tactile functaset (TactFunc) framework on reconstruction and downstream pose estimation task using data from two distinct tactile sensors' datasets: Bubble and Gelslim. We use the dataset first presented in \cite{rodriguez2024touch2touch} where images of both sensory modalities are recorded together with object poses and labels. 

The purpose of evaluating reconstruction is to demonstrate that despite the significant (several orders of magnitude) compression of our approach, the resulting reconstructions are high-quality, particularly when compared to baselines. The purpose of downstream task evaluation is to demonstrate that the representation enables simple models that generalize effectively across sensors. For reconstruction and inference tasks over our TactFunc, we explore the point estimate approach (Sec.~III C) when comparing to a baselines and demonstrate the use of SGLD (Sec.~III C) on the downstream task to show our models ability to sample from the posterior distribution of the latent modulation vector. We implement our models using the Jax Library~\cite{jax2018github}, drawing inspiration from~\cite{dupont2022datafunctadatapoint}. Our implementation leverages Haiku~\cite{haiku2020github} and Jaxline~\cite{jaxline} to facilitate the training process.

As baselines, we compare our method to (i) a ResNet-18~\cite{he2015deepresiduallearningimage}, (ii) a variational autoencoder (VAE)~\cite{kingma2022autoencodingvariationalbayes}, and (iii) T3 \cite{zhao2024transferable}. In more detail, we train a ResNet-18~\cite{he2015deepresiduallearningimage} that takes raw tactile signals as input and output the in-hand object pose vectors. The Resnet model does not perform any reconstruction, and is intended to evaluate downstream task success using raw sensor data. The VAE model first encodes the tactile signals to a 512-dimension latent vector space. It then reconstructs the signals using a symmetric decoder as well as predicting the 3D pose vectors from the latent code. The VAE evaluates how a standard compression technique performs both on reconstruction and downstream task using compression. The T3 model is a recent tactile representation based on the ViT architecture \cite{dosovitskiy2020image} where a trunk transformer learns a shared representation across a number of tactile sensor, each of which has its own encoder. For all baseline models, we first resize the raw tactile signals to a size of (224, 224) for consistency.

\subsection{Creating the Tactile Functaset}
We combine the Soft Bubble and Gelslim sensor datasets from~\cite{rodriguez2024touch2touch}, whose dataset contains paired tactile signals with corresponding object poses from 12 different tools. There are 32264 data data points in total. We use 29038 (90\%) for training and the remaining 3226 for testing. Given the combined dataset, we first train a latent modulated SIREN model that is similar to~\cite{dupont2022datafunctadatapoint} as the trunk model. The model's input is the normalized difference between two tactile signals: one captured when the tactile sensor contacts the object and another when the sensor is not in contact with anything. With the weights of this trunk model fixed as an initialization, we then use 512-dimension latent vectors as modulations and fit them to each data point iteratively. For each data point, we optimize an initial latent modulation with 3 gradient steps and save it as the functa. As described in~\cite{dupont2022datafunctadatapoint}, 3 steps are enough for both training the trunk model and fitting modulations. 

\begin{figure}[t]
    \centering
    \includegraphics[width=\columnwidth]{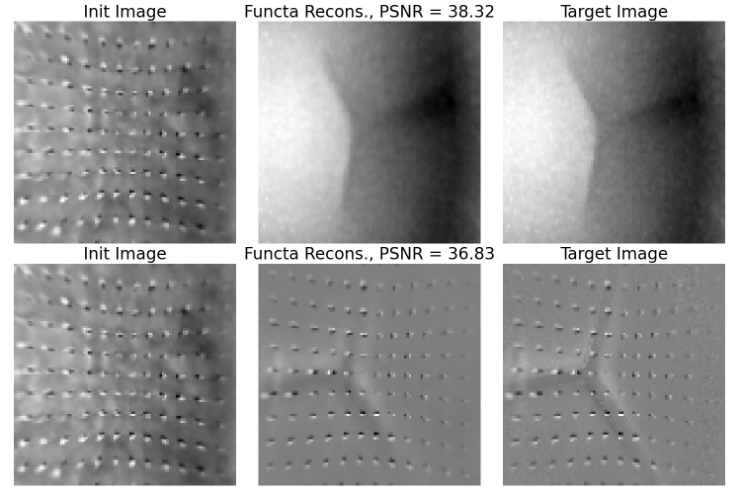}
    \caption{Comparison of three key stages in the image reconstruction process: the initial meta-learned state, the reconstructed image through gradient descent steps, and the desired outcome image.}
    \label{fig:tactfunc_recon}
\end{figure}

\begin{figure*}[t]
    \centering
    \begin{subfigure}[t]{0.4\linewidth}
        \includegraphics[width=\textwidth]{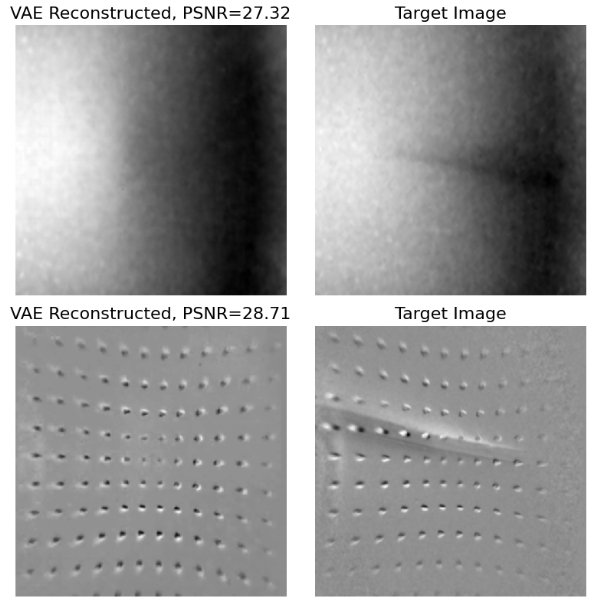}
    \end{subfigure}
    \hfill
    \begin{subfigure}[t]{0.48\linewidth}
        \includegraphics[width=\textwidth]{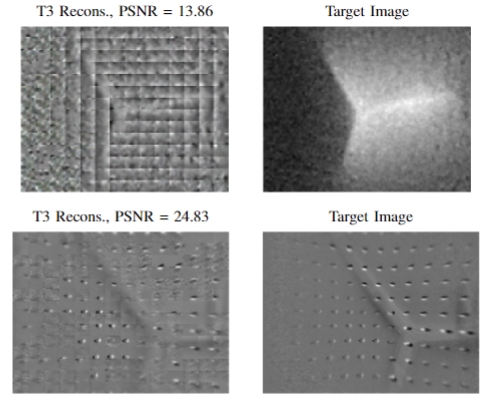}
    \end{subfigure}
    \caption{Visualization of VAE and T3 reconstructed images versus their target images.}
    \label{fig:image_comparison}
\end{figure*}

\subsection{Tactile Reconstruction}
To evaluate reconstruction quality, we employ standard visual metrics for image generation to compare our TactFunc's reconstructed signals against the ground truth signals obtained from the Soft Bubble and Gelslim sensors. We select Peak Signal to Noise Ratio (PSNR) as our visual metric. It exhibits high sensitivity to spatial positioning and has been extensively adopted in various touch generation studies, such as~\cite{yang2023generatingvisualscenestouch} and~\cite{dou2024tactileaugmentedradiancefields}. To reconstruct using TactFunc, we can simply merge the functa with the trunk model and pass the coordinates to it. As shown in Fig.~\ref{fig:tactfunc_recon}, the initial image (Init Image) shows the output of the trunk model merged with an empty initial latent modulation. Given two different target images with one from Soft Bubble and the other one from Gelslim, the Functa Reconstructed images show the direct output of the trunk model merged with corresponding tactile functa in the test set. Fig.~~\ref{fig:image_comparison} shows the reconstruction from the baseline models were we note the significant error. 
Tab.~I shows the Peak Signal-to-Noise Ratio (PSNR) of the reconstruction for our method and baselines. We note that despite having the smallest representation size (only requiring the storage of the base network and the modulations), our method produces the highest quality reconstructions.

\begin{table}[t]
\centering
\resizebox{\columnwidth}{!}{
\begin{tabular}{lclcc}
\toprule

\textbf{Approach} & \multicolumn{1}{l|}{\textbf{Rep. Size}} & \multicolumn{1}{l|}{\textbf{Split}} & \textbf{Reconstruction} & \textbf{Pose Estimation} \\
\midrule

\rowcolor[HTML]{D7FADB} 
\cellcolor[HTML]{D7FADB}                           & \cellcolor[HTML]{D7FADB}                   & Train & - & $\textbf{1.19}\times\textbf{10}^{-5}$ \\
\rowcolor[HTML]{D7FADB} 
\multirow{-2}{*}{\cellcolor[HTML]{D7FADB}ResNet-18} & \multirow{-2}{*}{\cellcolor[HTML]{D7FADB}107563 MB} & Test  & - & $\textbf{6.03}\times\textbf{10}^{-5}$ \\

\rowcolor[HTML]{FCE1C9} 
\cellcolor[HTML]{FCE1C9}                            & \cellcolor[HTML]{FCE1C9}                   & Train & 26.18  & $1.88\times10^{-2}$ \\
\rowcolor[HTML]{FCE1C9} 
\multirow{-2}{*}{\cellcolor[HTML]{FCE1C9}VAE}       & \multirow{-2}{*}{\cellcolor[HTML]{FCE1C9}224 MB} & Test  & 26.15  & $1.91\times10^{-2}$ \\

\rowcolor[HTML]{C2C2F0} 
\cellcolor[HTML]{C2C2F0}                            & \cellcolor[HTML]{C2C2F0}                   & Train & 19.24  & $8.23\times10^{-4}$ \\  
\rowcolor[HTML]{C2C2F0} 
\multirow{-2}{*}{\cellcolor[HTML]{C2C2F0}T3}       & \multirow{-2}{*}{\cellcolor[HTML]{C2C2F0}213 MB} & Test  & 19.18  & $1.22\times10^{-3}$ \\

\rowcolor[HTML]{D0F0F2} 
\cellcolor[HTML]{D0F0F2}                            & \cellcolor[HTML]{D0F0F2}                   & Train & \textbf{37.96}  & $1.43\times10^{-4}$ \\
\rowcolor[HTML]{D0F0F2} 
\multirow{-2}{*}{\cellcolor[HTML]{D0F0F2}TactFunc}  & \multirow{-2}{*}{\cellcolor[HTML]{D0F0F2}\textbf{83 MB}} & Test  & \textbf{37.89}  & $3.00\times10^{-4}$ \\

\bottomrule
\end{tabular}
}
\caption{The representation (Rep.) size encompasses two components: models and the dataset specific to each approach. For dataset, ResNet-18 uses the raw tactile image dataset, VAE creates a collection of latent vectors given its encoder, while for TactFunc, it comprises the Functaset. We evaluate the reconstruction quality of VAE and TactFunc using the Peak Signal-to-Noise Ratio (PSNR) measured in decibels {[}dB{]}. For the downstream pose estimation task, we employ MSE loss to assess the performance across all approaches: $\epsilon = \sqrt{\delta x^2 + \delta y^2 + \delta \theta^2}$ where the linear directions are in m$^2$ and the angle is in Radians.}
\label{results}
\end{table}

\subsection{Downstream Task -- In-hand Pose Estimation}

For downstream model learning, we focus on predicting in-hand object pose using our datasets labels. Building on our TactFunc framework, we implement a 3 linear fully connected MLP that consists of three hidden linear layers, each with a width of 512 and ReLU activations. The network takes as input our functa, represented by 512-dimensional modulations, and produces the object pose represented as a vector in $\mathrm{SE}(2)$.

Tab.~\ref{results} shows the resulting pose estimation errors for our method and the baselines. The Resnet model outperforms other models since it specializes in predicting the pose only (no reconstruction). However, because it does no reconstruction, the entire tactile image dataset must be maintained, hence the very large representation size. Our method is competitive with the Resnet model while requireing a significantly smaller size as only the trunk network and modulations need to be stored. Interestingly, the VAE model performs 3 orders of magnitude worse in pose estimation as well as having a larger representation size. The T3 model is closer to the Resnet and TactFunc models, though it is still an order of magnitude worse in testing than TactFunc while having a larger size. This is likely due to the fact that the proposed method to train T3 \cite{zhao2024transferable} is to learn reconstruction first, then fine-tune for downstream object classification and pose estimation. This fine-tuning comes at the cost of reconstruction performance.

\begin{figure}[t]
    \centering
    \includegraphics[width=\columnwidth]{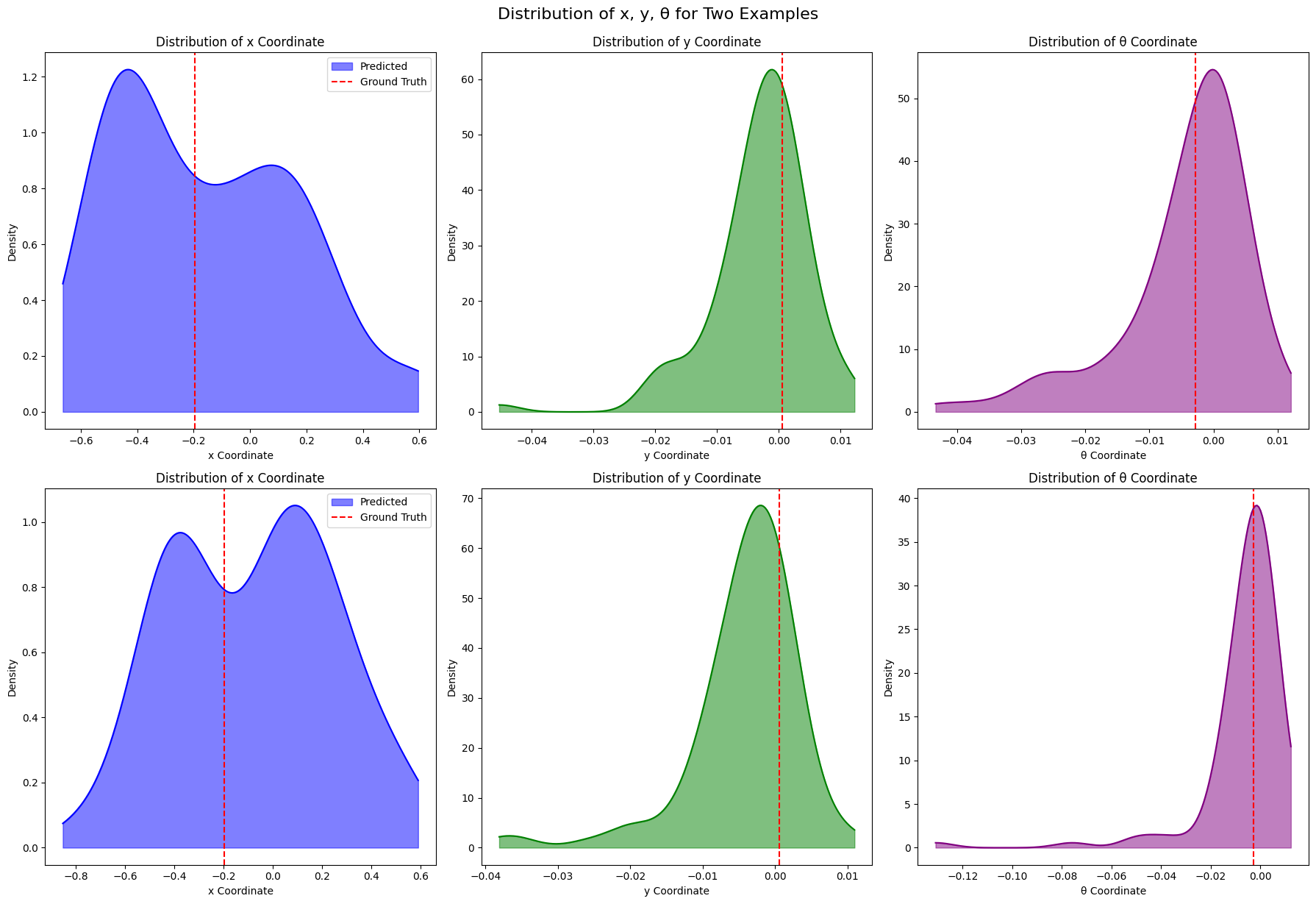}
    \caption{Example distributions for pose estimation, illustrating uncertainty quantification with the functaset representation. The distributions are most dense over the ground truth which indicates the fidelity of the model.}
    \label{fig:uncertainty}
\end{figure}

\noindent\textbf{Uncertainty Estimates for In-hand Pose Estimation:} In addition to point estimates of the grasped object pose, our model is capable of producing distributions using SGLD as outlined in Sec. III. C). In this section, we illustrate this capability with several visualizations of these distributions, Fig.~\ref{fig:uncertainty}. We use $N=100$ samples and zero mean Gaussian noise with $\sigma = 0.01$ and run SGLD for $k=3$ iterations. The fidelity of the model is due to the fact that the  majority of the distribution densities are centered about the ground-truth values. In addition to this accuracy, the uncertainty can be used in downstream tasks.
We emphasize that these examples are from unseen objects and inferred on the test set.


\section{DISCUSSION \& LIMITATIONS}

In recent years, robotics research has increasingly gravitated toward the collection and utilization of large-scale datasets, followed by training large-scale models to solve complex tasks. While this approach has yielded significant breakthroughs, it also brings substantial challenges in terms of storage and data processing. High-dimensional sensory data, such as tactile signals, often require extensive computational resources to process and store, making it difficult to scale robotic systems in practice.

Our work seeks to mitigate these issues specifically for tactile sensing. The key insight presented in this paper is that tactile signatures can be efficiently compressed using neural implicit functions. By encoding tactile data as compact, continuous function representations, we significantly reduce the storage burden, while maintaining sufficient information to enable downstream tasks. Rather than relying on raw data streams from sensors, we use these compressed neural implicit functions as inputs to downstream models. This not only reduces the computational complexity but also enables more efficient probabilistic inference and model-based reasoning.

However, our method comes with certain limitations. First, there is a substantial upfront computational cost associated with distilling raw tactile sensory data into the functaset. While this process compresses data and facilitates easier downstream task learning, it can be resource-intensive, particularly when processing large-scale datasets. Additionally, our method would benefit from advances in online learning. Currently, our approach relies on batch processing, and as a result, updating the model with new data can lead to costly retraining or catastrophic forgetting. Progress in online learning techniques, where models can adapt continuously without losing previously learned information, will greatly enhance the applicability of our framework in real-time and dynamic environments. 

Another consideration is that implicit functions provide a highly compact and flexible representation of tactile data, but they might lack the interpretability that other methods provide. The representations learned by these models are often black-boxes, which could be a limitation in safety-critical robotics applications where understanding the system's behavior is essential.
\bibliographystyle{ieeetr}
\bibliography{reference.bib}


\end{document}